\documentclass[10pt,conference,a4paper]{IEEEtran}
\usepackage{cite}
\ifCLASSINFOpdf
  \usepackage[pdftex]{graphicx}
\else
\fi
\usepackage{booktabs}
\begin{document}
\title{Comparison of deep learning and hand crafted features for mining simulation data}

% \author{\IEEEauthorblockN{Theodoros Georgiou}
% \IEEEauthorblockA{Leiden Institute of \\Advanced Computer Science\\
% Leiden, the Netherlands\\
% Email: t.k.georgiou@liacs.leidenuniv.nl}
% \and
% \IEEEauthorblockN{Sebastian Schmitt}
% \IEEEauthorblockA{Twentieth Century Fox\\
% Email: homer@thesimpsons.com}
% \and
% \IEEEauthorblockN{James Kirk\\ and Montgomery Scott}
% \IEEEauthorblockA{Starfleet Academy\\
% San Francisco, California 96678--2391\\
% Telephone: (800) 555--1212\\


\author{\IEEEauthorblockN{Theodoros Georgiou\IEEEauthorrefmark{1},
Sebastian Schmitt\IEEEauthorrefmark{2},
Thomas B\"ack\IEEEauthorrefmark{1},
Nan Pu\IEEEauthorrefmark{1},
Wei Chen\IEEEauthorrefmark{1} and
Michael Lew\IEEEauthorrefmark{1}}
\IEEEauthorblockA{\IEEEauthorrefmark{1}Leiden University\\
Leiden Institute of Advanced Computer Science,
Leiden, the Netherlands\\ Email: t.k.georgiou@liacs.leidenuniv.nl}
\IEEEauthorblockA{\IEEEauthorrefmark{2}Honda Research Institute Europe GmbH\\
Offenbach, Germany}}


\maketitle


\begin{abstract}

Computational Fluid Dynamics (CFD) simulations are a very important tool for many industrial applications, such as aerodynamic optimization of engineering designs like cars shapes, airplanes parts etc. The output of such simulations, in particular the calculated flow fields, are usually very complex and hard to interpret for realistic three-dimensional real-world applications, especially if time-dependent simulations are investigated. Automated data analysis methods are warranted but a non-trivial obstacle is given by the very large dimensionality of the data. A flow field typically consists of six measurement values for each point of the computational grid in 3D space and time (velocity vector values, turbulent kinetic energy, pressure and viscosity). In this paper we address the task of extracting meaningful results in an automated manner from such high dimensional data sets. We propose deep learning methods which are capable of processing such data and which can be trained to solve relevant tasks on simulation data, i.e. predicting drag and lift forces applied on an airfoil. We also propose an adaptation of the classical hand crafted features known from computer vision to address the same problem and compare a large variety of descriptors and detectors. Finally, we compile a large dataset of 2D simulations of the flow field around airfoils which contains 16000 flow fields with which we tested and compared approaches. Our results show that the deep learning-based methods, as well as hand crafted feature based approaches, are well-capable to accurately describe the content of the CFD simulation output on the proposed dataset.

% Computational Fluid Dynamics (CFD) simulations are very important for a plethora of industrial applications, such as optimization of aerodynamic properties of different engineering designs, e.g. cars, airplanes etc.. The output of these simulations can become very complex and hard to interpret due to their high dimensionality (3D space + time dependence with more than six values per point in the space-time). There have been many works that try to automatically extract meaningful information from these kind of simulations. In this paper we (i) propose an adaptation of the classical hand crafted features from computer vision for the purpose of describing such content, (ii) adapt deep learning methods for this kind of data, (iii) construct a dataset of 2D simulations for the purpose of benchmarking methods and (iv) make a comparison of these methods in being able to accurately describe the content of CFD simulation output on the proposed dataset.

\end{abstract}


% \fi
\IEEEpeerreviewmaketitle

\section{Introduction}
\label{sec:introduction}
Computational Fluid Dynamics (CFD) simulations provide a relatively fast way to evaluate and optimize the performance of different engineering designs. For example, estimations for drag and lift forces of moving objects such as cars or airplanes, as well as tumble motion patterns in internal combustions engines can readily be extracted. The availability of such simulations as well as their high complexity, which renders them difficult to analyze, motivate the development of methods that can analyze them automatically. For example, \cite{georgiou2018learning} developed and applied deep learning techniques in order to analyze the simulation while taking into account all the information produced by the simulation. Most existing feature extraction techniques, developed for CFD simulations focus on visualization and not machine learning \cite{UnsteadyflowVisualSoA2011,flowVisualSoA2016}. Deep learning techniques require exhaustive datasets with a very large  number of data samples to produce reliable and generalizable performance. On the other hand, CFD simulations are computationally very expensive and large datasets are therefore hard to produce. Additionally, each data sample is substantially bigger, compared to typical input of deep learning pipelines, such as images and videos, due to the high dimensionality and the high spatio-temporal resolutions typical for engineering applications. These contradicting issues set the stage for the challenging research field considered in this work where deep learning methods are applied to CFD simulation data.\par
% In this work we adapt the well known Bog of Words (BoW) technique from computer Vision to the case of CFD simulations and compare its performance to deep learning techniques.\par
% Training deep learning models requires a large number training examples. CFD simulations are very computationally expensive and thus producing large and diverse enough datasets for training deep learning models is very expensive, and sometimes infeasible. On the other hand, hand crafted features do not require as much data. Thus, when dealing with such expensive CFD simulations an appealing solution is to use hand crafted features.\par
Hand crafted features are a well known topic in computer vision. A big variety of features has been proposed along with methods that utilize them for numerous applications. One of the most well known is the SIFT detector and descriptor \cite{lowe2004distinctive}. After its success, a number of different descriptors and detectors were proposed, in order to improve performance, to be more efficient, or both. A very well known one is the SURF \cite{bay2006surf} descriptor and detector as well as a number of different binary descriptors, such as ORB \cite{rublee2011orb}, BRIEF\cite{calonder2010brief}, BRISK \cite{leutenegger2011brisk} and FREAK \cite{alahi2012freak}. In this work we utilize those features in the context of CFD simulation output. To the best of our knowledge we are the first to do so.\par
In recent years, deep learning approaches are outperforming the more traditional approach of feature extraction and description in most applications. Nonetheless, for some applications hand crafted features are better suited, for example when hardware availability is limited. Since, running complex CFD simulations takes a very long time, e.g. the simulation of a combustion process in an engine can take more than a month to compute, handcrafted features, which do not rely on a data-heavy training procedure, might still be a viable option. In order to validate that hypothesis we test a number of different detectors and descriptors on their ability to represent different flow fields and to discriminate between them. Moreover, we implement and test several deep learning approaches and compare them to the aforementioned hand crafted features. Finally, we compile a new dataset consisting of 16K 2D flow fields of the air around airfoils and utilize it as our benchmark platform.\par
The rest of the paper is organized as following. Section \ref{sec:relatedWork} discusses the relevant work to this paper, section \ref{sec:handCrafted} we describe the hand crafted features tested, as well as the implementation details. Section \ref{sec:deepLearning} discusses the deep learning techniques used, section \ref{sec:dataset} describes the dataset and benchmark developed for the purpose of this comparison. In section \ref{sec:experiments} we report and comment on our experimental results and finally, in section \ref{sec:conclusion} we draw our conclusions.\par

\section{Related Work}
\label{sec:relatedWork}

Most existing approaches for feature extraction on CFD simulations focus on visualization \cite{UnsteadyflowVisualSoA2011,flowVisualSoA2016}. Moreover, they focus specifically on the vector field and neglect other information, such as pressure and turbulent viscosity. The only work we are aware of that applied hand crafted features for machine learning on CFD simulation output, collects a number of streamlines, i.e. theoretical particle path in a vector field, with which they described each example \cite{graening2015flow}.\par
% Although efficient and effective it is applicable only to specific problems whilst it is very dependent on the designs that are  simulated.\par
Hand crafted features from computer vision, such as SIFT \cite{lowe2004distinctive}, SURF \cite{bay2006surf} and ORB \cite{rublee2011orb}, have been studied in much detail and applied to a plethora of applications \cite{schonberger2017comparative,liu2017local}. Nonetheless, to the best of our knowledge, we are the first to apply these features on CFD simulation output. There have been many comparisons between the detectors and descriptors \cite{schonberger2017comparative,liu2017local,ozaydin2019comparison,salahat2017recent,li2015survey}, but since there have been no applications of them on CFD simulation output, there is also no performance comparison.\par
In recent years, deep learning has become a mainstream approach for processing a number of different data types \cite{Georgiou2019,guo2018review} and especially data types that show some spatial or temporal relationship within each example, making the convolution a very effective operator. Since each example on the CFD simulation output can show both spatial and temporal relationships, deep learning and more specifically deep convolutional neural networks are an obvious choice for applying machine learning. As such there have been several applications of D-CNNs on CFD data \cite{Guo:2016:CNN:2939672.2939738,ling2016reynolds,georgiou2018learning,wu2020deep}. Most of these, though, focus on either predicting the flow itself \cite{Guo:2016:CNN:2939672.2939738,wu2020deep}, and thus substituting the simulation or substituting parts of the simulation with a deep learning predictor \cite{ling2016reynolds}. The work in \cite{georgiou2018learning} is the first we are aware of that applies deep learning on the output of CFD simulations in order to extract features from it, and thus the most relevant to our work. In this work we adapt some of the methods introduced in \cite{georgiou2018learning} to the 2D case. The work of work of \cite{marcos2017rotation} proposed a deep learning architecture for processing vector fields. A large part of the output of the CFD simulation is the velocity vector field. Thus we also test a combination of it with the approaches in \cite{georgiou2018learning}. Finally, we perform a comparison between the deep learning and traditional approaches, based on hand crafted features.

\section{Hand Crafted Features}
\label{sec:handCrafted}

Hand crafted feature detection and description is a very well studied subject in computer vision \cite{schonberger2017comparative,liu2017local,ozaydin2019comparison,salahat2017recent,li2015survey}. These features have been utilized for many applications, such as image classification \cite{nowak2006sampling}, image retrieval \cite{sivic2003video}, object detection \cite{girshick2011object} and scene semantic segmentation. Depending on the application the utilization strategy may vary. For example, when used for object detection separate descriptors are matched and aligned \cite{lowe2004distinctive}. In applications such as image classification or image retrieval, it is more useful to create a global description of the image. A common approach towards a global image description is to create the so called Bag of Words (BoW) model \cite{sivic2003video}. For the purpose of this comparison we also target a global description of the flow with which we aim to predict the drag and lift forces applied on an airfoil.\par
In the literature a multitude of approaches exists, both for detecting and describing visual features. Moreover there have been numerous studies that compare them for many applications. In order to limit the search we choose the most popular and best performing detectors and descriptors, according to the studies we found \cite{schonberger2017comparative,liu2017local,ozaydin2019comparison,salahat2017recent,li2015survey}. CFD simulation output is very peculiar in comparison to natural imagery. Each example consists of multiple data modalities and the values of each modality usually change smoothly. Thus, detectors created for natural images with many corners and abrupt changes don't detect many points. This raises some issues since some detector-descriptor combinations completely fail to find any features for some examples. In these cases it would be completely impossible to create a global description. Thus, these detectors are neglected from our study. Moreover, we tested extracting features from a regular grid instead of detected keypoints, as this strategy has proven to be superior in several applications that require global image description \cite{nowak2006sampling,wang2009evaluation}. The combinations tested, as well as whether they were successful in producing a global description for all examples are given in Tab. \ref{tab:handCraftedCombinations}. Besides these detectors and descriptors the CenSure \cite{agrawal2008censure}, Harris-Laplace \cite{mikolajczyk2004scale} detectors as well as the BRIEF \cite{calonder2010brief} and FREAK \cite{alahi2012freak} descriptors were tested, but we did not manage to get comparable results with any combination and thus are not included this comparison.\par

\begin{table}[!t]
\renewcommand{\arraystretch}{1.3}
\caption{Combination of detectors and descriptors tested. "SD" signifies that the combination is used with the single dictionary approach, "MD" signifies that the combination is used with the multiple dictionaries approach, "x" signifies that the combination didn't manage to produce results and "-" signifies that the combination was not tested.}
\label{tab:handCraftedCombinations}
\centering
\begin{tabular}{lcccc}
\toprule
  & SIFT \cite{lowe2004distinctive} & SURF \cite{bay2006surf} & ORB \cite{rublee2011orb} & AGAST \cite{mair2010adaptive} \\
\midrule
SIFT \cite{lowe2004distinctive} & x,MD    & -     & -     & SD,MD \\
SURF \cite{bay2006surf} & -     & SD,MD & -     & SD,MD \\
ORB  \cite{rublee2011orb} & -     & -     & SD,MD & SD,MD \\
BRISK \cite{leutenegger2011brisk} & -     & -     & -     & SD,x \\
\bottomrule
\end{tabular}
\end{table}

\begin{figure*}[h]
\centering
\includegraphics[width=\textwidth]{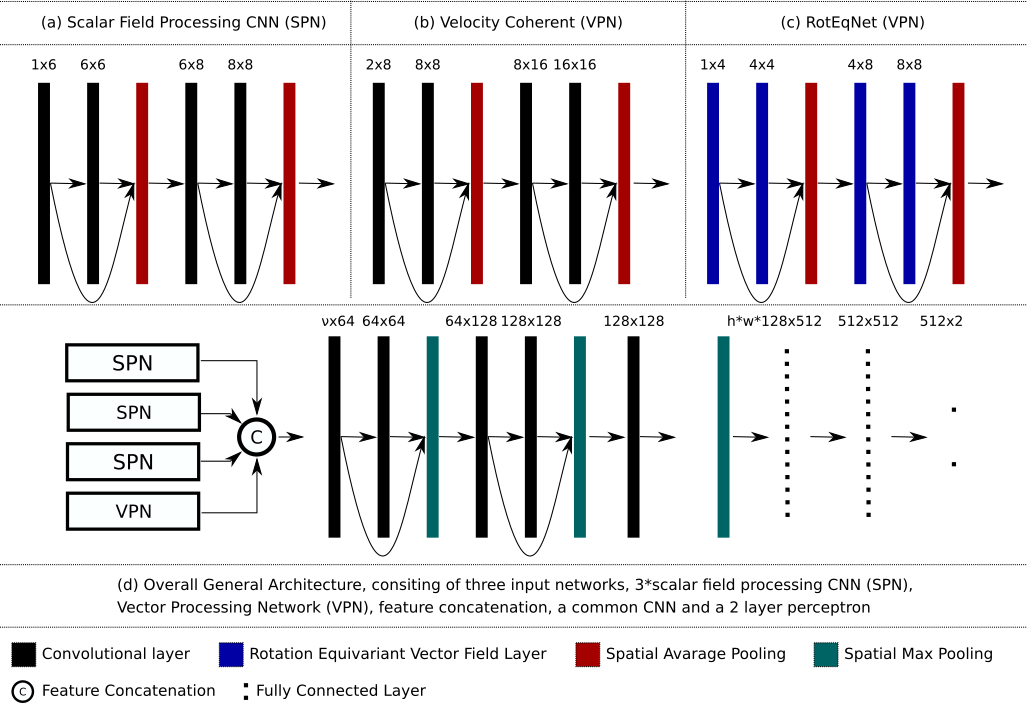}
\caption{CNN architecture. (a) Is the implementation of the input network processing the scalar fields (SPN). The same network is applied separately to each input field. (b) The Vector Processing Network (VPN) for the VC strategy. (c) The VPN network for the RotEqNet strategy. (d) The complete architecture. The VPN depends on the chosen strategy. It is Either (b), (c) or in the case of DS 2*SPN. The numbers above each layer are the number of input and output filters respectively. In the case of RotEqNet, the number of channels is double the number of filters (both input and output) because each filter is represented by the magnitude and the angle of a vector. h,w are the height and width of the input feature maps.}
\label{fig:cnnArchitectures}
\end{figure*}
One of the peculiarities of our data, compared with traditional imagery, is the number of modalities. Overall there are five modalities, i.e. two for the velocity vector field and three for pressure, turbulent viscosity and a viscosity related field from the turbulence modeling, i.e. $\tilde{\nu}$ from SpalartAllmaras RAS model \cite{openFOAM}. According to our study \cite{lai2011large,silberman2011indoor,Georgiou2019}, the most common approach for combining information from multiple modalities is to create process each modality separately, e.g. create a dictionary for each modality and concatenate the per-modality representation to construct the final representation \cite{lai2011large} or perform classification based on each modality and then average the results \cite{silberman2011indoor}. In this study we also explore a slightly different approach. We extract features from all modalities and filter out common features by discarding the ones that have intersection over union ratio above 0.9. Then, for each detected feature, we describe them for all modalities and concatenate the representations. Finally, we construct one common dictionary from the concatenated features. In order to differentiate between the different strategies, we call the common approach "Multiple Dictionaries" (MD) and the second one "Single Dictionary" (SD). Finally, we also tested dense feature extraction, which we denote as "DE". \par
After acquiring the global description of each example, we train a Random Forest (RF) regressor to predict the drag and lift forces. We utilize the OpenCV \cite{opencv_library} implementations of the detectors and descriptors from the Python API. The dictionaries for real valued descriptors is built using an approximate K-Means clustering, whilst for binary descriptors we used the K-Majority algorithm with Hamming distance as the distance metric. The approximation of our K-Means, pre-computes all the pairwise distances of data points and sets as cluster center the closest point to the actual cluster center. Thus, we don't need to compute the distances of all points to all centers in each step since we already have them. We utilize the scikit-learn package's \cite{scikit-learn} implementation of the RF with default parameters, while the clustering algorithms are implemented by us.\par

\section{Deep Learning Approaches}
\label{sec:deepLearning}

\subsection{Methodology}
As mentioned in section \ref{sec:relatedWork}, deep learning has already been applied to the same subject. Georgiou et. al. \cite{georgiou2018learning} proposed and tested three different strategies for processing the multi-modal data produced by the simulations. Based on their results, we pick the Velocity Coherent (VC) and Direction Specific (DS) approaches as our baseline models and adjust them for the 2D case. Specifically, both approaches use one network with one input channel to process the scalar fields (applied to each one separately). The VC approach utilizes a network with two input channels to process the velocity vector field whilst the DS uses two networks with one input channel, one for each direction of the velocity. These input processing networks are comprised by four convolutional layers. Their output is concatenated and then passed to another CNN for further processing. The structure ends with three fully connected layers the last of which is performing the drag and lift regression. Besides reducing the dimensionality to two, we also added skip connections every two layers, since it proved to increase the performance of the networks.\par
Moreover, as mentioned in section \ref{sec:relatedWork}, we define a third approach based on the RotEqNet, proposed by Marcos et. al.\cite{marcos2017rotation}. In their work, they propose an architecture tailored for vectorized data. We utilize this approach by using the RotEqNet as a substitute of the velocity processing network of the VC and DS approaches. The output of each layer is a vector field for each filter, resulting in double number of channels compared to a plain CNN with the same number of filters. Moreover, since the input consists of vector field feature maps, the number of trainable parameters is given by: $F_h \cdot F_w \cdot 2 \cdot c_i \cdot c_o$, where $F_h,F_w$ are the filter height and width respectively and $c_i$, $c_o$ the input and output number of channels. The three different input pipelines are shown in Fig. \ref{fig:cnnArchitectures}, top. After the processing of each modality, the feature maps are concatenated and further processed by a common CNN of five layers. Finally, three fully connected layers perform the regression.\par

\subsection{Implementation and training details}
For all strategies, the input processing networks are four layers deep and the network applied after the feature concatenation consists of five convolutional layers. Spatial average pooling operations are performed every two convolutional layers in the input processing networks, whilst spatial max pooling is performed every two convolutional layers in the network after the feature concatenation. All convolutional kernels have spatial dimensions $5\times5$ and are followed by batch normalization \cite{ioffe2015batch} and a leaky ReLU activation function \cite{maas2013rectifier} where $\alpha=0.1$. The number of nodes per layer is given in Fig. \ref{fig:cnnArchitectures}. The prediction network consists of a max pooling operation, followed by three fully connected layers, with 512, 512 and 2 nodes respectively. The third fully connected layer is tasked to predict the drag and lift forces. All networks are trained with Adam optimizer \cite{kingma2014adam}, with the default parameters and a batch size of 200, for 36K iterations. Our implementation is done using Tensorflow 1.13.1 \cite{tensorflow2015-whitepaper} and all experiments ran on NVIDIA GTX 1080Ti graphics cards.

\section{Dataset}
\label{sec:dataset}

The aim of this paper is to compare the performance of deep learning based and more traditional hand crafted feature based approaches, for mining CFD simulation output. Due to the much larger variety of hand crafted features for 2D imagery, as well as the high computational demand of deep learning methods on high dimensional data, we pick the 2D simulation domain as the setting for our benchmark, since it exhibits many of the characteristics that exist in the 3D domain, such as a large number of modalities and a velocity vector field, that satisfy, to a certain extent, the Navier-stokes equations.\par
\begin{figure}[h]
\centering
\includegraphics[width=\linewidth]{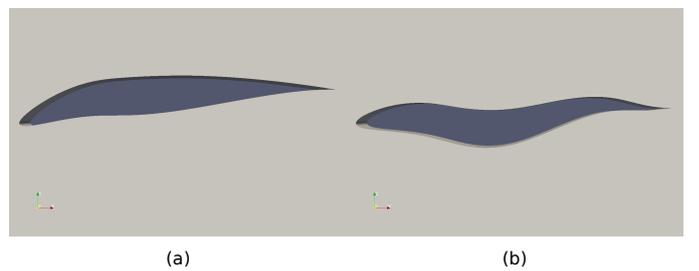}
\caption{Example airfoil shapes. (a) Baseline model. (b) Randomly deformed shape.}
\label{fig:shapes}
\end{figure}

To the best of our knowledge there is no 2D dataset we can utilize as our benchmark. Consequently, we propose our own dataset. We focus on the standard airfoil example. In order to create a large dataset with as big variety as possible we follow a similar approach to \cite{georgiou2018learning}. First, we get a baseline airfoil shape (Fig. \ref{fig:shapes}) and apply random deformations on it. Then, given intake air from the left of the simulation domain, we simulate the air around the airfoil using Reynolds-Average Simulator (RAS) implemented in OpenFOAM-v5 \cite{openFOAM}. The output of the simulation consists of the velocity vector field, the pressure field, turbulent viscosity, and the drag and lift forces applied on the airfoil. Overall we create 2K shapes and simulate for 8 different angles of attack per shape, resulting in 16K simulations. 15K are chosen for training at random and the remaining 1K are using as a test set.\par

\begin{figure}[h]
\centering
\includegraphics[width=\linewidth]{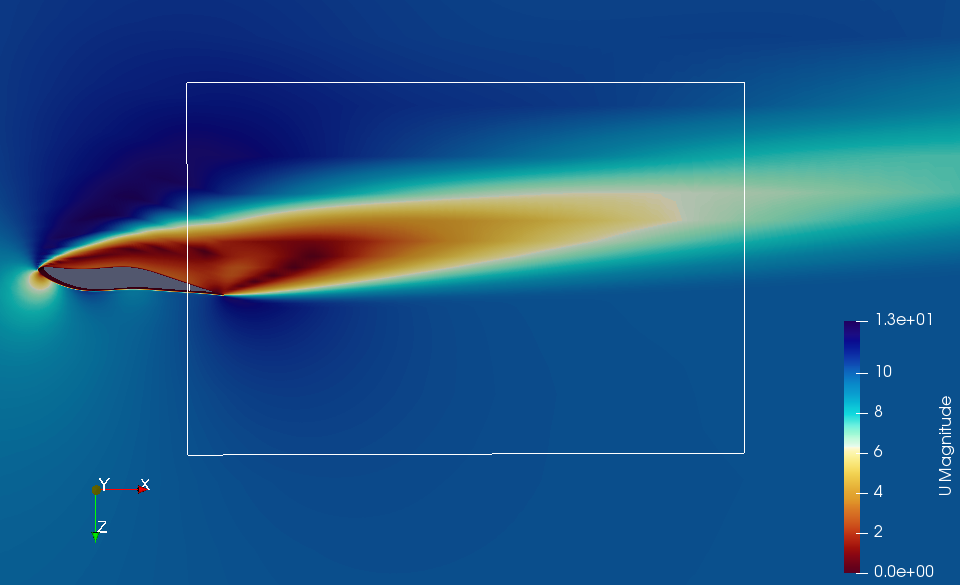}
\caption{Example simulation. The square behind the airfoil is the window used as an example for our pipeline.}
\label{fig:crop}
\end{figure}

The aim of this work is to perform data mining and pattern recognition on the flow fields. Thus, we want to discard any information that relates to the shape of the airfoil. Consequently, we crop a window behind the airfoil (see Fig. \ref{fig:crop}) and take the flow field in this region for further processing. The simulation is performed on an unstructured mesh. In order to bring the data to a format that hand crafted features and CNNs can be easily applied to, we interpolate the values on a regular grid with resolution $192\times128$. Finally, to evaluate whether the defined methods are able to extract meaningful information, they aim is to predict the drag and lift forces applied on the airfoil. The dataset is publicly available on Zenodo \footnote{https://zenodo.org/record/4077323\#.X4QeI3VfhhE}.\par

\section{Experiments}
\label{sec:experiments}

In order to asses whether the features are capable of encoding relevant information, we used them to predict drag and lift force coefficients of each airfoil. We then compare the predicted values with the actual simulation results and quantify the performance using the root-mean squared error (RMSE) as metric.\par
With our first experiment we aim at identifying the optimal number of clusters which are used to construct the dictionaries. We tested dictionary sizes of 512, 1024 and 2048 for the SD approaches and summarize the results in Table \ref{tab:SD_dictionarySizes}.

\begin{table}[!t]
\renewcommand{\arraystretch}{1.3}
\caption{Regression performance of the SD approach with varying dictionary sizes, measured by RMSE. The best performing method per column is highlighted with italics, and the overall best for each evaluation measure (Drag or Lift) is highlighted with bold.}
\label{tab:SD_dictionarySizes}
\centering
\begin{tabular}{llrrr|rrr}
\toprule
Detector & Descriptor & \multicolumn{3}{c}{Drag (*1e-3)} & \multicolumn{3}{c}{Lift (*1e-2)}\\
\midrule
 & & 512 & 1024 & 2048 & 512 & 1024 & 2048 \\
\midrule
 SURF  & SURF  & 10.17 & 9.37 & 9.57 & 5.6 & 5.59 & 6.02 \\
 ORB   & ORB   & \textit{7.88} & 8.22 & 7.77 & \textit{4.74} & \textit{4.1} & \textbf{3.87} \\
 AGAST & SIFT  & 8 & \textit{7.83} & \textbf{7.51} & 5.25 & 4.72 & 4.88 \\
 AGAST & SURF  & 10.06 & 9.89 & 10.49 & 7.73 & 7.39 & 8.64 \\
 AGAST & ORB   & 7.93 & 7.86 & 7.84 & 5.8 & 5.56 & 5.69 \\
 AGAST & BRISK & 8.67 & 8.42 & 8.95 & 7.95 & 7.42 & 7.32 \\
\bottomrule
\end{tabular}
\end{table}

For the MD approach, we need to set the number of clusters for each modality, Most of the detectors detected extremely low number of points for some of the modalities, e.g. pressure, and thus we were unable to use large dictionary sizes. As a result the dictionary sizes for the two velocity directions as well as pressure are set to 32 and 64. For the rest of the modalities we tested 512 and 1024. The results are given in Table \ref{tab:MD_dictionarySizes}.\par

\begin{table}[!t]
\renewcommand{\arraystretch}{1.3}
\caption{Regression performance of the MD approach with varying dictionary sizes, measured by RMSE. The best performing method per column is highlighted with italics, and the overall best for each evaluation measure (Drag or Lift) is highlighted with bold.}
\label{tab:MD_dictionarySizes}
\centering
\begin{tabular}{llrr|rr}
\toprule
Detector & Descriptor & \multicolumn{2}{c}{Drag (*1e-3)} & \multicolumn{2}{c}{Lift (*1e-2)}\\
\midrule
 & & 32-512 & 64-1024 & 32-512 & 64-1024 \\
\midrule
 SIFT  & SIFT  & 19.55 & 16.59 & 15.16 & 14.11 \\
 SURF  & SURF  & 11.67  & 9.52 & 7.97 & 9.56 \\
 ORB   & ORB   & 8.52  & 8.84 & \textit{4.6}  & \textbf{4.52} \\
 AGAST & SIFT  & \textbf{7.22}  & \textit{7.68} & 6.68 & 7.3 \\
 AGAST & SURF  & 9.99 & 9.92 & 11.98 & 10.24 \\
 AGAST & ORB   & 9.76 & 9.26 & 7.15 & 6.93 \\
\bottomrule
\end{tabular}
\end{table}

Looking at Tables \ref{tab:SD_dictionarySizes} and \ref{tab:MD_dictionarySizes} we can identify a few trends. Overall, the ORB-ORB combination produces the best performance for predicting the lift force while being a close second on the lift prediction performance. The AGAST-SIFT combination similarly has the highest performance in predicting drag forces and a close second on predicting lift forces. Regarding the modality aggregation strategy, for most detector-descriptor combinations, the SD approach outperformed the MD approach, with only exceptions the AGAST-SIFT on drag prediction and the AGAST-ORB on lift prediction.\par
For dense sampling we extract features in 4 different scales, i.e. \{12, 16, 24, 32\} pixels. The step size for each scale is the same number of pixels as the size of the scale. Regarding modality aggregation we tried both approaches, i.e. SD and MD. Due to time limitations we only experimented with the best performing descriptors, according to our previous experiments, namely ORB and SIFT. For the MD approach we tested dictionary sizes of \{256, 512\}, for each modality, resulting in a 1280 and 2560 global description size, respectively. For the SD approach we tested the same dictionary sizes with the detector approach. The results can be seen in Tables \ref{tab:DE_SD_results} and \ref{tab:DE_MD_results}.\par
The results show that dense sampling outperforms the detection mechanism in terms of global description in most cases, similar to what is found for other computer vision tasks. This is not the case only for the combination of ORB descripton in the context of drag prediction. Moreover, in contrast to the use of detectors, the MD approach performs better than the SD approach. The dense description approach increased the quality of the results for the SIFT by a significant margin, rendering it the highest performing method for both tasks. In contrast, the performance increase is not found with the ORB descriptor, where the performance even dropped by a significant margin.\par

\begin{table}[!t]
\renewcommand{\arraystretch}{1.3}
\caption{Regression performance, measured by RMSE, of the DE-SD approach with varying dictionary sizes and modality aggregation strategies.}
\label{tab:DE_SD_results}
\centering
\begin{tabular}{lrrr|rrr}
\toprule
Descriptor & \multicolumn{3}{c}{Drag (*1e-3)} & \multicolumn{3}{c}{Lift (*1e-2)}\\
\midrule
 & 512 & 1024 & 2048 & 512 & 1024 & 2048 \\
\midrule
 ORB & 10.46 & 9.22 & 9.26 & 3.34 & 3.58 & 3.94 \\
 SIFT & \textit{7.03} & \textit{8.08} & \textbf{6.59} & \textit{2.68} & \textit{3.11} & \textbf{2.65} \\
 \bottomrule
\end{tabular}
\end{table}

\begin{table}[!t]
\renewcommand{\arraystretch}{1.3}
\caption{Regression performance, measured by RMSE, of the DE-MD approach with varying dictionary sizes and modality aggregation strategies.}
\label{tab:DE_MD_results}
\centering
\begin{tabular}{lrr|rr}
\toprule
Descriptor & \multicolumn{2}{c}{Drag (*1e-3)} & \multicolumn{2}{c}{Lift (*1e-2)}\\
\midrule
& 256 & 512 & 256 & 512 \\
\midrule
 ORB & 21.14 & 38.78 & 10.74 & 16.63 \\
 SIFT & \textbf{6.28} & \textit{9.09} & \textbf{2.33} & \textit{2.87} \\
 \bottomrule
\end{tabular}
\end{table}

Table \ref{tab:DL_results} shows the performance achieved by the deep learning approaches. Comparing Tables \ref{tab:DE_MD_results} and \ref{tab:DL_results}, deep learning approaches outperform hand crafted feature approaches in all benchmarks. Particularly, all deep learning approaches perform better than the DE-SIFT-MD, i.e. the best hand crafted feature based approach, for both drag and lift prediction. For a more thorough comparison relevant in practice, it needs to be stated that hand crafted feature approaches have less computational complexity. The DE-SIFT-MD approach with dictionary sizes of 256 per modality takes 7.5K seconds to extract features from the training set, cluster them and train the random forest on two Intel(R) Xeon(R) CPU E5-2699. At the same time the VC approach takes around 11.1K seconds to be trained on the same machine running on an NVIDIA GTX 1080Ti. Applying these approaches to large scale CFD simulations, with 4 physical dimensions and multiple modalities, where the data complexity is much larger, the high computational demand of deep learning approaches might render them infeasible, making the hand crafted feature based approaches an appealing alternative.\par
For a further comparison between the features produced by deep learning and the hand crafted features, we extract the output of the last fully connected layer of our best network (VC) and train a RF regressor. The result is given in the last row of Table \ref{tab:DL_results}, depicted as VC-RF. It is apparent that the use of CNNs as feature extractor and a random forest regressor to perform the given task achieves much higher performance than the equivalent neural network solution, or the use of hand crafted feature based description with an RF regressor.\par

\begin{table}[!t]
\renewcommand{\arraystretch}{1.3}
\caption{Regression performance, measured by RMSE, of the Deep learning approaches.}
\label{tab:DL_results}
\centering
\begin{tabular}{lrr}
\toprule
Approach & Drag (*1e-3) & Lift (*1e-2)\\
\midrule
 VC & 5.32 & 2.18 \\
 DS & 5.53 & 2.25 \\
 RotEqNet & 5.22 & 2.2 \\
 VC-RF & \textbf{2.83} & \textbf{0.92} \\
 \bottomrule
\end{tabular}
\end{table}

\begin{figure}[h]
\centering
\includegraphics[width=\linewidth]{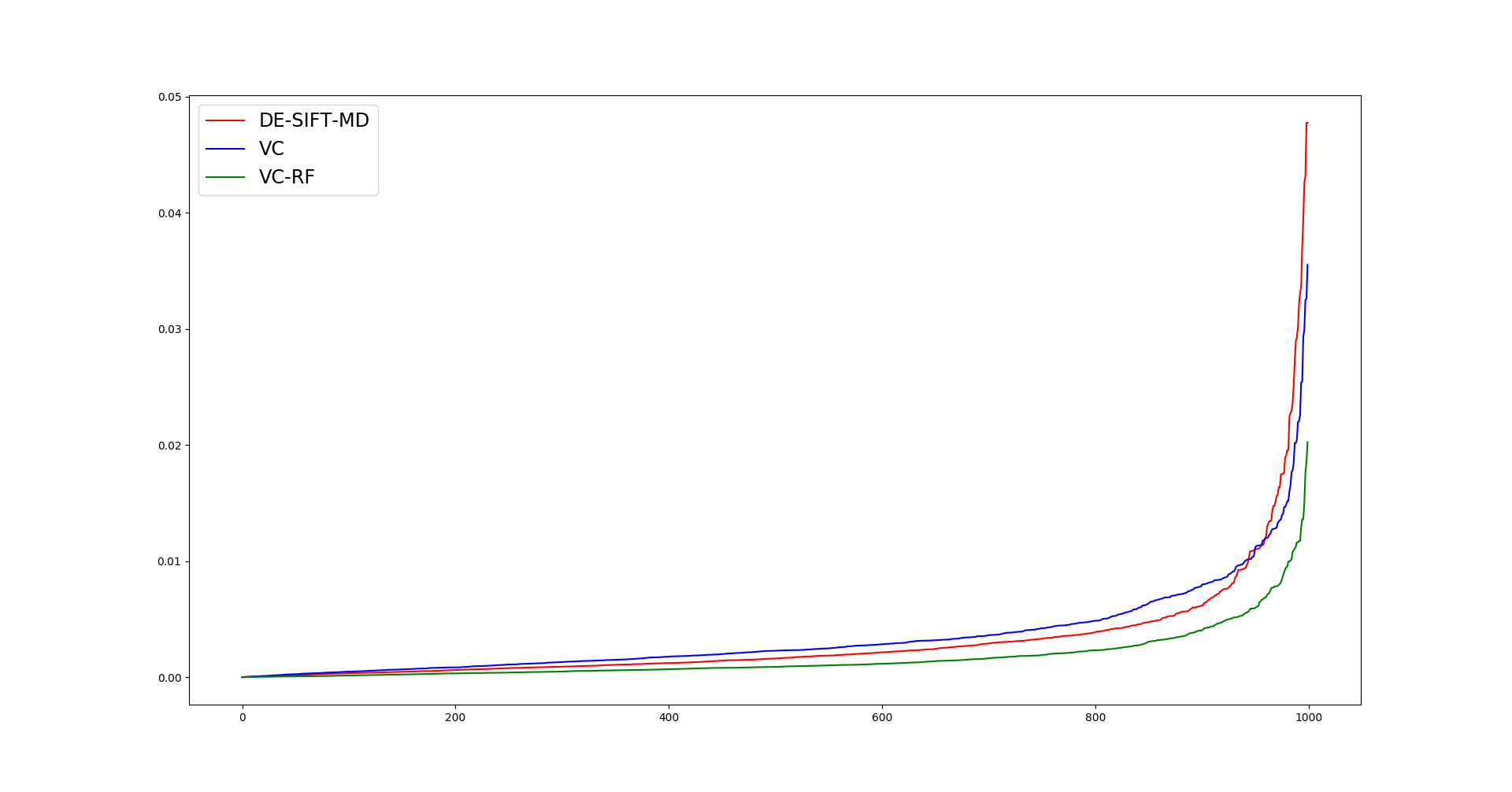}
\includegraphics[width=\linewidth]{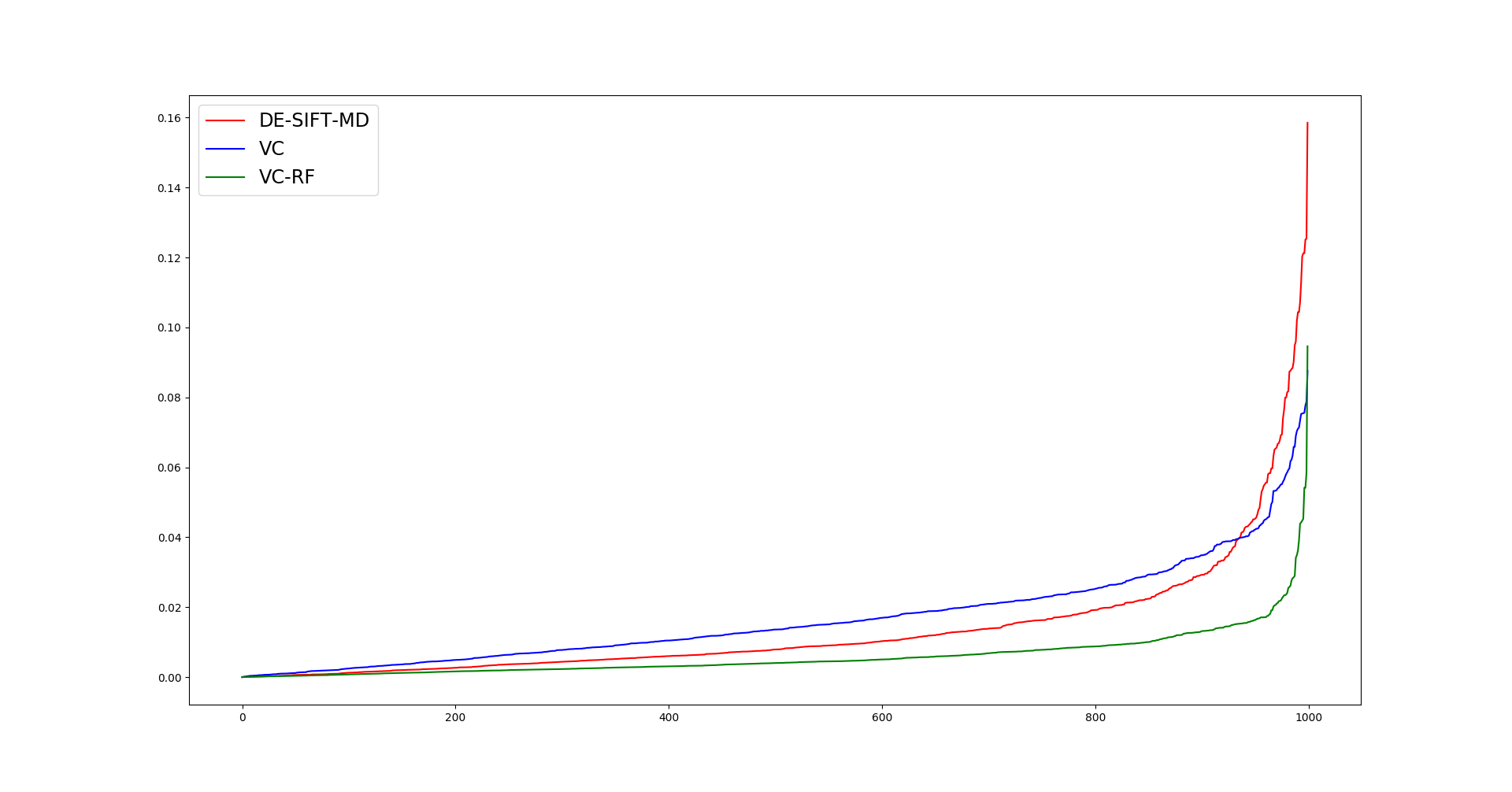}
\caption{Sorted absolute drag (top) and lift (bot) regression error per test example. The Y-axis is the absolute error, whilst the X-axis is the index of the test examples (after sorting based on the absolute error).}
\label{fig:regressionError}
\end{figure}

In order to get a more informative image of the performance of our methods, we also calculate the $R^2$ values for the three top performing methods, i.e. the DE-SIFT-MD, the VC and the VC-RF and show them in Table \ref{tab:rSquared}. Moreover, we plot the sorted absolute errors, per test example, for both lift and drag forces in Figure \ref{fig:regressionError}. There is a qualitative difference between the performance of the VC and the DE-SIFT-MD approaches. The VC approach has much lower $R^2$ values whilst it achieves lower RMSE. This can be explained by the two figures. Although overall the DE-SIFT-MD approach has lower absolute errors, the error of the extreme cases become much more severe than in the case of the VC approach. Depending on the behavior one needs from a system, a different approach would be preferable. Finally, in both drag and lift prediction, the VC-RF approach manages to produce much better results in all measures tested, RMSE, $R^2$ as well as overall better absolute error curves.\par

\begin{table}[!t]
\renewcommand{\arraystretch}{1.3}
\caption{Regression performance, measured by $R^2$, of the three best performing approaches. The best performing method is highlighted with bold.}
\label{tab:rSquared}
\centering
\begin{tabular}{lrr}
\toprule
Approach & Drag & Lift \\
\midrule
 DE-SIFT-MD & 0.965 & 0.987 \\
 VC & 0.917 & 0.949 \\
 VC-RF & \textbf{0.981} & \textbf{0.994} \\
 \bottomrule
\end{tabular}
\end{table}

One of the motivations of this work was to assess the usability of hand-crafted features for CFD simulation output and compare them to deep learning solutions for situations with small training datasets. In order to do this, we ran the best performing configurations, from both deep learning and hand-crafted features with varying training set sizes and the same validation and test sets. Our training set sizes are \{1K, 2K, 4K, 8K, 14K\}. The performance for drag and lift forces are shown in Tables \ref{tab:perfPertsSize_drag} and \ref{tab:perfPertsSize_lift} respectively. It is apparent, that even with small training set sizes the deep learning approaches outperform the hand-crafted features. Surprisingly, the difference between the performances becomes larger as the training set size reduces, showing that deep learning approaches are more capable of maintaining relatively good performance even with small training set sizes.\par

\begin{table}[h]
\renewcommand{\arraystretch}{1.3}
\caption{Drag regression performance, measured by RMSE, of the DE-SIFT-MD and VC with varying training set sizes.}
\label{tab:perfPertsSize_drag}
\centering
\begin{tabular}{lrrrrr}
\toprule
 & 1K & 2K & 4K & 8K & 14K \\
\midrule
 VC & \textbf{11.7} & \textbf{8.39} & \textbf{6.81} & \textbf{6.05} & \textbf{5.32} \\
 DE-SIFT-MD & 21.26 & 19.26 & 14.56 & 12.19 & 6.27 \\
 \bottomrule
\end{tabular}
\end{table}

\begin{table}[h]
\renewcommand{\arraystretch}{1.3}
\caption{Lift regression performance, measured by RMSE, of the DE-SIFT-MD and VC with varying training set sizes.}
\label{tab:perfPertsSize_lift}
\centering
\begin{tabular}{lrrrrr}
\toprule
 & 1K & 2K & 4K & 8K & 14K \\
\midrule
 VC & \textbf{4} & \textbf{3.05} & \textbf{2.71} & \textbf{2.3} & \textbf{2.18} \\
 DE-SIFT-MD & 10.3 & 7.36 & 6.16 & 5.19 & 2.33 \\
 \bottomrule
\end{tabular}
\end{table}

\section{Conclusion}
\label{sec:conclusion}

In this work we evaluate the performance of hand crafted features such as SIFT \cite{lowe2004distinctive}, SURF \cite{bay2006surf} and ORB \cite{rublee2011orb} on their ability to efficiently describe CFD simulation output and compare them to a number of deep learning approaches. CFD simulation output can be very complex and large, compared to standard computer vision application examples, such as 2D and 3D imagery. Moreover, creating these examples can even take months making the generation of enough examples for deep learning approaches infeasible. On the other hand, a complete working pipeline based on hand crafted features might not require as many examples, since the basic features are predefined and not learned from the data. Due to the large variety of detectors and descriptors that exist in 2D, as well as the lower computational complexity of 2D CFD simulations compared to 3D \& 4D, we create a dataset of 2D CFD simulations and use it as our benchmark platform.\par
Overall we tested 4 detectors and 6 descriptors as well as dense sampling. Moreover, we tried 2 different approaches for combining different data modalities, namely a multiple dictionary (MD) approach and a single dictionary (SD) approach. Their difference lies in the concatenation step. Specifically, in the SD approach we concatenate the low level features, before the dictionary construction, whilst in the MD approach we concatenate the information after we create the dictionaries. According to our experiments, for the drag and lift prediction tasks, dense sampling combined with SIFT descriptor produces the best results of all hand crafted feature based approaches by a large margin. Moreover, we identify a contradiction. In most cases, the SD approach outperforms the MD approach, but with dense sampling and SIFT features, which is the best combination tested, we see the opposite behavior.\par
We also implemented and tested a number of deep learning approaches. They are adapted approaches from data mining on 3D simulation data \cite{georgiou2018learning}. We also combine them with the work of Marcos et al. \cite{marcos2017rotation}, which is designed for vector fields. Deep learning methods outperformed the hand crafted feature based approaches in the benchmarks tested. Moreover, our experiments showed that using the neural networks as feature extractors whilst a random forest regressor to perform the task produces higher performance than the pure deep-NN counterpart. Comparing different regression characteristics, like $R^2$ and the absolute error curves, we see a qualitative difference between the DE-SIFT-MD and the VC approaches. The DE-SIFT-MD produces more accurate predictions on most test instances but also gets much higher maximum error than the VC approach. Moreover, the $R^2$ performance of the DE-SIFT-MD is higher than that of the VC, even thought the the RMSE of VC is lower than that of DE-SIFT-MD. Thus, depending on the requirements of an application, different method would be preferable.\par
We compared the deep learning methods to the hand crafted features in terms of efficiency. As expected, the hand crafted feature approach is more efficient as it managed to construct the global description of all examples as well as train the RF regressor in 67\% of the time it took to train the best performing CNN.\par
The 2D airfoil example is considered a very important benchmark for CFD simulations. Nonetheless, it is much simpler than many industrial application CFD simulations. As such, we are able to get a much higher number of examples for our dataset. In order to get an intuition on how our methods would generalize to the more realistic case with much smaller datasets, we perform an experiment and train our models on much smaller training set sizes while testing on the same test set. Our results show that deep learning approaches are much more capable of getting a high performance with the drop of the training set size, than the hand-crafted feature based approaches. We speculate that the reason behind this is the necessity of a large number of descriptor required to build generalizable enough dictionaries.\par

\section*{Acknowledgment}

This work is part of the research program DAMIOSO with project number 628.006.002, which is partly financed by the Netherlands Organization for Scientific Research (NWO) and partly by Honda Research Institute-Europe (GmbH).




\bibliographystyle{IEEEtran}
% \bibliography{handCraftedOnCFD}


% \end{thebibliography}

\end{document}